%% file: paper.tex
\documentclass{article}

\usepackage{microtype}
\usepackage{graphicx}
\usepackage{array}
\usepackage{multirow}
\usepackage{subcaption}
\usepackage{algorithm}
\usepackage{algorithmic}
\usepackage{flushend}
\usepackage{booktabs} 
\usepackage{resizegather}
\usepackage{hyperref}

\usepackage[accepted]{icml2024}

\usepackage{amsmath}
\usepackage{amssymb}
\usepackage{mathtools}
\usepackage{amsthm}

\usepackage[capitalize,noabbrev]{cleveref}

\theoremstyle{plain}

\theoremstyle{definition}

\theoremstyle{remark}

\usepackage{amsmath}
\usepackage{mathtools}
\usepackage{amsthm}
\usepackage{graphicx}
\usepackage{hyperref}
\usepackage{url}
\usepackage{caption}
\usepackage{subcaption}
\usepackage{bbm}
\usepackage{graphicx}
\usepackage{amsmath,amssymb,amsfonts,nccmath}
\graphicspath{ {./images/} }
\usepackage[textsize=tiny]{todonotes}

\icmltitlerunning{Overcoming Class Imbalance: 
Unified GNN Learning with Structural and Semantic Connectivity Representations}

\begin{document}

\twocolumn[
\icmltitle{Overcoming Class Imbalance: 
Unified GNN Learning with Structural and Semantic Connectivity Representations}



\icmlsetsymbol{equal}{*}

\begin{icmlauthorlist}
	\icmlauthor{Abdullah Alchihabi}{cu}
	\icmlauthor{Hao Yan}{cu}
	\icmlauthor{Yuhong Guo}{cu}
\end{icmlauthorlist}

\icmlaffiliation{cu}{School of Computer Science, Carleton University, Canada}


\icmlkeywords{GNN Learning, Class Imbalance}

\vskip 0.3in
]



\printAffiliationsAndNotice{}  

\begin{abstract}
Class imbalance is pervasive in real-world graph datasets, where the majority of annotated nodes 
belong to a small set of classes (majority classes), leaving many other
	classes (minority classes) with only a handful of labeled nodes. 
	Graph Neural Networks (GNNs) suffer from significant performance degradation 
	in the presence of class imbalance, 
exhibiting bias towards majority classes and struggling to generalize effectively on minority classes.
This limitation stems, in part, from the message passing process, leading GNNs to overfit to the limited neighborhood 
	of annotated nodes from minority classes and impeding the propagation of discriminative information throughout the entire graph.
In this paper, we introduce a novel Unified Graph Neural Network Learning (Uni-GNN) framework to tackle 
	class-imbalanced node classification.
The proposed framework seamlessly integrates both structural and semantic connectivity representations through 
semantic and structural node encoders. 
By combining these
connectivity types, Uni-GNN extends the propagation of node embeddings beyond immediate neighbors, 
	encompassing non-adjacent structural nodes and semantically similar nodes, enabling efficient diffusion 
	of discriminative information throughout the graph.
Moreover, to harness the potential of unlabeled nodes within the graph, we employ a balanced pseudo-label generation 
	mechanism that augments the pool of available labeled nodes from minority classes in the training set. 
Experimental results underscore the superior performance of our proposed Uni-GNN framework compared to 
	state-of-the-art class-imbalanced graph learning baselines across
	multiple benchmark datasets.
\end{abstract}

\section{Introduction}

Graph Neural Networks (GNNs) have exhibited significant success in addressing the node classification task \cite{kipf2017semisupervised, hamilton2017inductive, velickovic2018graph} across diverse application domains from molecular biology \cite{hao2020asgn} to fraud detection \cite{zhang2021fraudre}.
The efficacy of GNNs has been particularly notable when applied to balanced annotated datasets, 
where all classes have a similar number of labeled training instances.
The performance of 
GNNs experiences a notable degradation when confronted with an increasingly imbalanced class distribution in the available training instances \cite{yun2022lte4g}. This decline in performance materializes as a bias towards the majority classes, which possess a considerable number of labeled instances, resulting in a challenge to generalize effectively over minority classes that have fewer labeled instances \cite{park2021graphens, yan2023unreal}. The root of this issue lies in 
GNNs' reliance on message passing to disseminate information across the graph. Specifically, when the number of labeled nodes for a particular class is limited, GNNs struggle to propagate discriminative information related to that class throughout the entire graph. This tendency leads to GNNs' overfitting to the confined neighborhood of labeled nodes belonging to minority classes \cite{tang2020investigating, yun2022lte4g, GraphSHA}. 
This is commonly denoted as the `under-reaching problem' \cite{sun2022position} or `neighborhood memorization' \cite{park2021graphens}.

Class-imbalanced real-world graph data are widespread, spanning various application domains such as 
the Internet of Things \cite{wang2022minority}, 
Fraud Detection \cite{zhang2021fraudre}, 
and Cognitive Diagnosis \cite{wang2023self}. Consequently, there is a critical need to develop 
GNN models that demonstrate robustness to class imbalance, avoiding biases towards majority classes while maintaining the ability to generalize effectively over minority classes.
Traditional methods addressing class imbalance, such as oversampling \cite{chawla2002smote} 
or re-weighting \cite{yuan2012sampling}, face limitations in the context of graph-structured data as they do not account for the inherent graph structure. Consequently, several approaches have been proposed to specifically tackle class imbalance within the realm of semi-supervised node classification.
Topology-aware re-weighting methods, which consider the graph connectivity when assigning weights to labeled nodes, such as TAM \cite{song2022tam}, have demonstrated improved performance compared to traditional re-weighting methods. However, these methods still exhibit limitations as they do not effectively address issues related to neighborhood memorization and the under-reaching problem.
Node oversampling methods, including ImGAGN \cite{qu2021imgagn}, GraphSMOTE \cite{zhao2021graphsmote}, and GraphENS \cite{park2021graphens}, generate new nodes and establish connections with the existing graph through various mechanisms. Despite their potential, these methods face an open challenge in determining the optimal way to connect synthesized nodes to the rest of the graph. Additionally, they often fall short in harnessing the untapped potential of the substantial number of unlabeled nodes present in the graph.

In this study, we present a novel Unified Graph Neural Network Learning (Uni-GNN) framework designed to tackle the challenges posed by class-imbalanced node classification tasks. Our proposed framework leverages both structural and semantic connectivity representations, specifically addressing the under-reaching and neighborhood memorization issues.
To achieve this, we construct a structural connectivity based on the input graph structure, complemented by a semantic connectivity derived from the similarity between node embeddings. Within each layer of the Uni-GNN framework, we establish a dedicated message passing layer for each type of connectivity. This allows for the propagation of node messages across both structural and semantic connectivity types, resulting in the acquisition of comprehensive structural and semantic representations.
The Uni-GNN framework's unique utilization of both structural and semantic connectivity empowers it to effectively 
extend the propagation of node embeddings beyond 
the standard neighborhood. 
This extension reaches non-direct 
 structural neighbors and semantically similar nodes, facilitating the efficient dissemination of discriminative information throughout the entire graph.
Moreover, to harness the potential of unlabeled nodes in the graph, we introduce a balanced pseudo-label generation method. This method strategically samples unlabeled nodes with confident predictions in a class-balanced manner, effectively increasing the number of labeled instances for minority classes.
Our experimental evaluations on multiple benchmark datasets underscore the superior performance of the proposed Uni-GNN framework compared to state-of-the-art Graph Neural Network methods designed to address class imbalance.

\section{Related Works}

\subsection{Class Imbalanced Learning}

Class-imbalanced learning methods generally fall into three categories: re-sampling, re-weighting, and data augmentation. 
Re-sampling involves adjusting the original imbalanced data distribution to a class-balanced one by either up-sampling minority class data or down-sampling majority class data \cite{wallace2011class, mahajan2018exploring}. 
Re-weighting methods assign smaller weights to majority instances and larger weights to minority instances, either in the loss function or at the logits level \cite{ren2020balanced, menon2020long}. 
Data augmentation methods either transfer knowledge from head classes to augment tail classes or apply augmentation techniques to generate additional data
for minority classes \cite{chawla2002smote, kim2020m2m, zhong2021improving, ahn2023cuda}.
Traditional oversampling methods do not consider the graph structure when generating synthesized samples, making them unsuitable for graph data. Self-training methods for semi-supervised class imbalanced learning address class imbalance by supplementing the labeled set with unlabeled nodes predicted to belong to the minority class \cite{wei2021crest}. 
However, GNNs' bias towards majority classes can introduce noise in pseudo-labels, necessitating the 
development of new methods to mitigate the bias under class imbalance.

\subsection{Class Imbalanced Graph Learning}

Traditional class imbalance methods assume independently distributed labeled instances, a condition not met in graph data where nodes influence each other through the graph structure. Existing class imbalanced graph learning can be categorized into three groups: re-weighting, over-sampling, and ensemble methods.
Re-weighting methods assign importance weights to labeled nodes based on their class labels while considering the graph structure. Topology-Aware Margin (TAM) adjusts node logits by incorporating local connectivity \cite{song2022tam}. However, re-weighting methods have limitations in addressing the under-reaching or neighborhood memorization problems \cite{park2021graphens, sun2022position}.
Graph over-sampling methods generate new nodes and connect them to the existing graph. 
GraphSMOTE uses SMOTE to synthesize minority nodes in the learned embedding space and connects them using an edge generator module \cite{zhao2021graphsmote}. ImGAGN employs a generative adversarial approach to balance the input graph by generating synthetic nodes with semantic and topological features \cite{qu2021imgagn}. GraphENS generates new nodes by mixing minority class nodes with others in the graph, using saliency-based node mixing for features and KL divergence for structure information \cite{park2021graphens}. However, these methods increase the graph size, leading to higher computational costs.
Long-Tail Experts for Graphs (LTE4G) is an ensemble method that divides training nodes based on class and degree into subsets, trains expert teacher models, and uses knowledge distillation to obtain student models \cite{yun2022lte4g}.
It however comes with a computational cost due to training multiple models.

\section{Method}

In the context of semi-supervised node classification with class imbalance, we consider a graph $G=(V,E)$, where $V$ represents the set of 
$N=|V|$ nodes, 
and $E$ denotes the set of edges within the graph. 
$E$ is commonly represented by an adjacency matrix $A$ of size $N\times N$. 
This matrix is assumed to be symmetric (i.e. for undirected graphs), 
and it may contain either weighted or binary values.
Each node in the graph is associated with a feature vector of size $D$,
while the feature vectors for all the $N$ nodes are organized into an input feature matrix $X\in \mathbb{R}^{N\times D}$. 
The set of nodes $V$ is partitioned into two distinct subsets: $V_{l}$, 
comprising labeled nodes, and $V_u$, encompassing unlabeled nodes.
The labeled nodes in $V_{l}$ are paired with class labels, and this information is encapsulated in a label indicator matrix $Y^{l} \in \{0,1\}^{N_{l \times C}}$. Here, $C$ signifies the number of classes, and $N_{l}$ is the number of labeled nodes. Further, $V_{l}$ can be subdivided into $C$ non-overlapping subsets, denoted as $\{V^{1}_{l},\cdots, V^{C}_{l}\}$, where each subset $V^{i}_{l}$ corresponds to the labeled nodes belonging to class $i$.
It is noteworthy that $V_{l}$ exhibits class imbalance, characterized by an imbalance ratio $\rho$. 
This ratio, defined as $\rho = \frac{\text{min}_{i}(|V^{i}_{l}|)}{\text{max}_{i}(|V^{i}_{l}|)}$, 
is considerably less than 1. Such class imbalance problem
introduces challenges and necessitates specialized techniques in the development of effective node classification models.

This section introduces the
proposed Unified Graph Neural Network Learning (Uni-GNN) framework designed specifically for addressing class-imbalanced 
semi-supervised
node classification tasks. 
The Uni-GNN framework integrates both structural and semantic connectivity to facilitate the learning of discriminative, unbiased node embeddings. 
Comprising two node encoders—structural and semantic—%
Uni-GNN ensures joint utilization of these two facets. 
The collaborative efforts of the structural and semantic encoders converge in the balanced node classifier, 
which effectively utilizes the merged output from both encoders to categorize nodes within the graph. Notably, 
a weighted loss function is employed to address the challenge of class-imbalanced nodes in the graph, ensuring that 
the classifier is robust and capable of handling imbalanced class distributions.
To harness the potential of unlabeled nodes, 
we introduce a balanced pseudo-label generation strategy. This strategy 
generates
class-balanced confident pseudo-labels for the unlabeled nodes, contributing to the overall robustness and effectiveness of the 
Uni-GNN framework. 
In the rest of this section, we delve into the 
details of both the Uni-GNN framework 
and the balanced pseudo-label generation strategy, 
as well as the synergies between them.

\subsection{Unified GNN Learning Framework}

\begin{figure}
    \centering \includegraphics[width=0.49\textwidth]{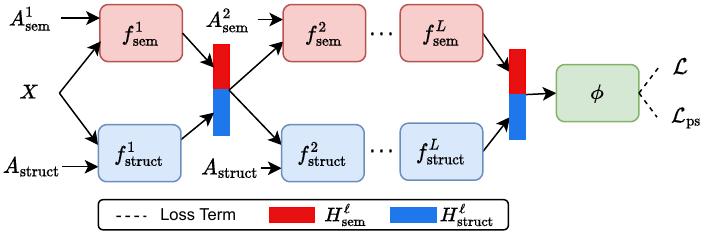}
    \caption{Overview of the proposed Unified GNN Learning framework. The structural ($f_{\text{struct}}$) and semantic ($f_{\text{sem}}$) node encoders leverage their respective connectivity matrices—structural ($A_{\text{struct}}$) and semantic ($\{A^{\ell}_{\text{sem}}\}_{\ell=1}^{L}$). The encoders share concatenated node embeddings—structural ($H^{\ell-1}_{\text{struct}}$) and semantic ($H^{\ell-1}_{\text{sem}}$)—at each message passing layer ($\ell$). The balanced node classifier ($\phi$) utilizes the final unified node embeddings ($H^{L}_{\text{struct}}||H^{L}_{\text{sem}}$) for both node classification and balanced pseudo-label generation.}
    \label{fig:diagram}
\vskip -0.2in
\end{figure}

The Unified GNN Learning (Uni-GNN) framework comprises three crucial components: the structural node encoder, the semantic node encoder and the balanced node classifier. 
The structural node encoder is dedicated to constructing a graph adjacency matrix founded on structural connectivity, facilitating the propagation of node embeddings beyond the immediate structural neighbors of nodes.
Concurrently, the semantic node encoder generates adjacency matrices based on semantic connectivity where it connects nodes to their semantically similar neighboring nodes, transcending structural distances. 
This facilitates the linking of nodes that are spatially distant in the graph but are from the same class, 
and enables message propagation between distantly located yet similar nodes, 
enriching the encoding process with learned node embeddings.
At every layer in both the structural and semantic encoders, 
we propagate the integrated structural and semantic embeddings of the nodes along the corresponding connectivity, 
alleviating bias induced by each single type of connectivity. 
Ultimately, a balanced node classifier utilizes the acquired node embeddings from both the structural and semantic encoders for node classification. 
The overall framework of Uni-GNN is illustrated in Figure \ref{fig:diagram}. 


\subsubsection{Structural Node Encoder}

The objective of the structural encoder is to learn an embedding of the nodes based on structural connectivity. 
Instead of directly using the input adjacency matrix $A$, 
we construct a
new structural connectivity-based graph adjacency matrix $A_{\text{struct}}$ 
that extends connections beyond the immediate neighbors in the input graph. 
This matrix is determined by the distances between pairs of nodes, measured in terms of the number of edges along the shortest path connecting the respective nodes, such that
\begin{equation}
 A_{\text{struct}}[i,j]=
         \begin{cases} 
        \frac{1}{\text{SPD}(i,j)} & \text{SPD}(i,j) \leq \alpha \\
        0 & \text{otherwise}\\
        \end{cases}
\label{eq:A_struct}
\end{equation}
where $\text{SPD}(.,.)$ is the shortest path distance function that measures the distance between pairs of input nodes in terms of the number of edges along the shortest path in the input graph $G$. 
The hyper-parameter $\alpha\geq 1$ governs the maximum length of the shortest path distance to be considered. 
In $A_{\text{struct}}$, 
edges connecting node pairs are assigned weights that are inversely proportional to the length of the shortest path between them. This design ensures that the propagated messages carry importance weights, scaling the messages based on the corresponding edge weights between connected nodes. 
The constructed structural connectivity enables us to directly propagate messages to nodes beyond a node's 
immediate structural neighbors in the original graph. 
This is beneficial for
expanding the influence of labeled minority nodes 
towards more distant neighboring nodes within the graph,
particularly when there are under-reaching problems 
induced by sparse connectivities in the original graph. 

The structural node encoder, denoted as $f_{\text{struct}}$, 
consists of $L$ message propagation layers. 
In each layer $\ell$ of the structural node encoder, denoted as $f^{\ell}_{\text{struct}}$, 
the node input features comprise the learned node embeddings, 
$H^{\ell-1}_{\text{struct}}$ and $H^{\ell-1}_{\text{sem}}$, 
from the previous layer of both the structural encoder 
and the semantic encoder, respectively.
As a consequence, the propagated messages encode both semantic and structural information facilitating the learning of more discriminative node embeddings. 
The constructed structural connectivity matrix $A_{\text{struct}}$ is employed as the adjacency matrix for message propagation within each layer. 
We employ the conventional Graph Convolution Network (GCN) \cite{kipf2017semisupervised} as our message-passing layer, 
given its simplicity, efficiency and ability to handle weighted graph structures, in the following manner:
\begin{equation}
\begin{aligned}
    &{H}_{\text{struct}}^{\ell} 
    =  f^{\ell}_{\text{struct}}( {H}_{\text{struct}}^{\ell-1} || {H}_{\text{sem}}^{\ell-1}  , A_{\text{struct}} )  \\
    &= \sigma\left(\hat{D}_{\text{struct}}^{-\frac{1}{2}} \hat{A}_{\text{struct}} \hat{D}_{\text{struct}}^{-\frac{1}{2}} ({H}_{\text{struct}}^{\ell-1} || {H}_{\text{sem}}^{\ell-1} ) W_{\text{struct}}^{\ell}  \right) 
\end{aligned}
\end{equation}
Here, $\sigma$ represents the non-linear activation function, 
$``||"$ denotes the feature concatenation operation, 
$W_{\text{struct}}^{\ell}$ is the matrix of learnable parameters for $f^{\ell}_{\text{struct}}$, 
$\hat{A}_{\text{struct}} = {A}_{\text{struct}} + I$ is the adjusted adjacency matrix with self-connections,
and $\hat{D}_{\text{struct}}$ is the diagonal node degree matrix of $\hat{A}_{\text{struct}}$ 
such that $\hat{D}_{\text{struct}}[i,i]=\sum_j \hat{A}_{\text{struct}}[i,j]$.
In the case of the first layer of $f_{\text{struct}}$, the node input features 
are solely represented by the input feature matrix $X$.

\subsubsection{Semantic Node Encoder}

The objective of the semantic node encoder is to learn node embeddings
based on the semantic connectivity. The semantic node encoder, denoted as $f_{\text{sem}}$, comprises $L$ message passing layers. In each layer $\ell$ of the semantic node encoder, represented by $f^{\ell}_{\text{sem}}$, a semantic-based graph adjacency matrix $A^{\ell}_{\text{sem}}$ is constructed based on the similarity between the embeddings of nodes from the previous layer of the semantic and structural node encoders, measured in terms of clustering assignments.
For each layer $\ell$, the following fine-grained node clustering is performed: 
\begin{equation}
    S^{\ell} = g( {H}_{\text{struct}}^{\ell-1} || {H}_{\text{sem}}^{\ell-1},K) 
\label{eq:clust_nodes}    
\end{equation}
which clusters all the graph nodes into $K$ ($K\gg C$) clusters. 
Here, $S^{\ell} \in \mathbb{R}^{N\times K}$ is the fine-grained clustering assignment matrix obtained from the clustering function $g$. 
The row $S^{\ell}[i]$ 
indicates the cluster to which node $i$ is assigned.
The fine-grained clustering function $g$ takes as input the concatenation of the structural and semantic node embeddings from layer $\ell-1$, along with the number of clusters $K$, 
and outputs the cluster assignments $S^\ell$. 
The clustering function $g$ is realized 
by performing K-means clustering to minimize the following least squares clustering loss:
\begin{equation}
\!\!\!	\mathcal{L}_{\text{clust}} = \sum_{i \in V} \sum_{k=1}^{K} S^\ell[i,k] 
	\left\lVert ({H}_{\text{struct}}^{\ell-1}[i] || {H}_{\text{sem}}^{\ell-1}[i])  - \boldsymbol{\mu}_k \right\rVert^2 
\label{eq:clust_loss}
\end{equation}
where $\boldsymbol{\mu}_k$ represents the mean vector 
for the cluster $k$, 
and $S^\ell[i,k]$ has a binary value (0 or 1) that indicates whether node $i$ is assigned to cluster $k$. 
Based on the fine-grained clustering assignment matrix $S^{\ell}$, the construction of the semantic connectivity-based graph adjacency 
matrix $A^{\ell}_{\text{sem}}$ is detailed as follows:
\begin{equation}
            A^{\ell}_{\text{sem}}[i,j]=
        \begin{cases}
        1 & \text{if }  S^{\ell}[i] = S^{\ell}[j] \\
        0 & \text{otherwise}. 
    \end{cases}
\label{eq:A_sem}
\end{equation}
In the construction of $A^{\ell}_{\text{sem}}$, nodes assigned to the same cluster are connected, 
establishing edges between them, while nodes assigned to different clusters are not connected, 
resulting in an adjacency matrix that encapsulates the semantic connectivity encoded 
within the fine-grained clusters.
This process enables message propagation among nodes that share semantic similarities in the graph, irrespective of their structural separation. This is instrumental in addressing the issue of under-reaching of minority nodes. 
Moreover, by constructing individual semantic adjacency matrix for each layer, 
we can prevent adherence to fixed local semantic clusters and enhance both robustness and adaptivity. 
The semantic connectivity matrix of the first layer of the semantic encoder, $A^{1}_{\text{sem}}$, 
is constructed based on the input features matrix $X$. 
Furthermore, to balance efficiency with stability during the training process, an update mechanism is introduced. 
Specifically, $A^{\ell}_{\text{sem}}$ is periodically updated by re-clustering the nodes based on 
the updated node embeddings along the training process. 
The update is performed at intervals of $\beta$ training iterations. 
This adaptive strategy ensures that the semantic connectivity information remains relevant and adapts to the evolving node embeddings during the training process.

Each layer $\ell$ of the semantic node encoder, 
$f^{\ell}_{\text{sem}}$, takes
the concatenation of node embeddings, 
$H^{\ell-1}_{\text{struct}}\| H^{\ell-1}_{\text{sem}}$,  
from the previous layer $\ell-1$ of both the structural encoder 
and the semantic encoder as input, 
aiming to gather richer information from both aspects, 
but propagates messages with the constructed semantic adjacency matrix $A^{\ell}_{\text{sem}}$. 
For the first layer of $f_{\text{sem}}$, the input node features are simply the input features matrix $X$. 
We opt for the conventional Graph Convolution Network (GCN) \cite{kipf2017semisupervised} as our message-passing layer again,
which is employed in the following manner:
\begin{equation}
\begin{aligned}
    &{H}_{\text{sem}}^{\ell} 
    =f^{\ell}_{\text{sem}}( {H}_{\text{struct}}^{\ell-1} || {H}_{\text{sem}}^{\ell-1}  , A^{\ell}_{\text{sem}})  \\
    &=\sigma\left(\hat{D}_{\text{sem}}^{-\frac{1}{2}} \hat{A}^{\ell}_{\text{sem}} \hat{D}_{\text{sem}}^{-\frac{1}{2}} ({H}_{\text{struct}}^{\ell-1} || {H}_{\text{sem}}^{\ell-1} ) W_{\text{sem}}^{\ell}  \right) 
\end{aligned}
\end{equation}
Here, $\sigma$ again represents the non-linear activation function; 
$W_{\text{sem}}^{\ell}$ is the matrix of learnable parameters for $f^{\ell}_{\text{sem}}$;
$\hat{A}_{\text{sem}}^{\ell} = A_{\text{sem}}^{\ell} + I$ is the adjusted adjacency matrix with self-connections; and 
the diagonal node degree matrix $\hat{D}_{\text{sem}}$ 
is computed as $\hat{D}_{\text{sem}}[i,i]=\sum_j \hat{A}_{\text{sem}}^{\ell}[i,j]$. 

\subsubsection{Balanced Node Classifier}

We define a balanced node classification function $\phi$, which classifies the nodes in the graph based on their structural and semantic embeddings learned by the Structural Encoder and Semantic Encoder respectively. In particular, the balanced node classification function takes as input the output of the 
$L$-th layers of the structural and semantic node encoders, denoted as ${H}_{\text{struct}}^{L}$ and ${H}_{\text{sem}}^{L}$, respectively: 
\begin{equation}
        P = \phi({H}_{\text{struct}}^{L} || {H}_{\text{sem}}^{L}) 
\end{equation}
where $P \in \mathbb{R}^{N\times C}$ 
is the predicted class probability matrix of all the nodes in the graph.
Given the class imbalance in the set of labeled nodes $V_{l}$, 
the node classification function is trained to minimize the following weighted node classification loss
on the labeled nodes: 
\begin{equation}
        \mathcal{L} = 
        \sum\nolimits_{i \in V_{l} } \omega_{c_i} \ell_{\text{ce}}( P_i , Y^l_i  ). 
\label{eq:sup_loss}
\end{equation}
Here, $\ell_{\text{ce}}$ denotes the standard cross-entropy loss function. 
For a given node $i$, $P_i$ represents its predicted class probability vector, 
and $Y^l_i$ is the true label indicator vector if $i$ is a labeled node.
The weight $\omega_{c_i}$ associated with each node $i$
is introduced to balance the contribution of data from different classes
in the supervised training loss.
It gives different weights to 
nodes from different classes. 
In particular, the class balance weight $\omega_{c_i}$ is calculated as follows:
\begin{equation}
        \omega_{c_i}= \frac{1}{|V_l^{c_i}|}
\end{equation}
where $c_i$ denotes the class index of node $i$, such that $Y^l[i, c_i]=1$; 
and $|V_l^{c_i}|$ is the size of class $c_i$ in the labeled nodes---
i.e., the number of labeled nodes $V_l^{c_i}$ from class $c_i$. 
Since $\omega_{c_i}$ is inversely proportional to the corresponding class size, 
it enforces that
larger weights are assigned to nodes from 
minority classes with fewer labeled instances 
in the supervised node classification loss,  
while smaller weights are assigned to nodes from majority classes with abundant labeled nodes. 
Specifically, through the incorporation of this class weighting mechanism, each class contributes equally to the supervised loss function, irrespective of the quantity of labeled nodes associated with it within the training set,
thereby promoting balanced learning across different classes.

\begin{table*}[t]
\caption{The overall performance (standard deviation within brackets) on Cora, CiteSeer and PubMed datasets with 2 different numbers of minority classes (3 and 5 on Cora and CiteSeer, 1 and 2 on Pubmed) and 2 imbalance ratios (10\% and 5\%). OOM indicates out-of-memory.} 
\setlength{\tabcolsep}{3.0pt}

\resizebox{\textwidth}{!}{
\begin{tabular}{ll|lll|lll|lll|lll}
\hline
& \# Min. Class & \multicolumn{6}{l|}{3}  & \multicolumn{6}{l}{5}           \\
\hline
& $\rho$  & \multicolumn{3}{l|}{10\%}    & \multicolumn{3}{l|}{5\%}  & \multicolumn{3}{l|}{10\%}                          & \multicolumn{3}{l}{5\%}                           \\
\hline
&                & bAcc.          & Macro-F1        & G-Means        & bAcc.          & Macro-F1       & G-Means        & bAcc.          & Macro-F1       & G-Means        & bAcc.           & Macro-F1        & G-Means        \\
\hline
\multirow{9}{*}{\rotatebox{90}{Cora}}     & GCN         & $68.8_{(4.0)}$ & $68.8_{(4.0)}$  & $80.8_{(2.6)}$ & $60.0_{(0.4)}$ & $56.6_{(0.7)}$ & $74.8_{(0.3)}$ & $64.9_{(6.2)}$ & $64.7_{(5.7)}$ & $78.1_{(4.2)}$ & $55.1_{(2.6)}$  & $51.4_{(2.4)}$  & $71.4_{(1.8)}$ \\
    & Over-sampling  & $65.6_{(4.3)}$ & $63.4_{(5.6)}$  & $78.6_{(2.9)}$ & $59.0_{(2.2)}$ & $53.9_{(2.6)}$ & $74.2_{(1.5)}$ & $58.9_{(5.6)}$ & $56.9_{(7.6)}$ & $74.0_{(3.9)}$ & $49.1_{(4.1)}$  & $45.6_{(5.4)}$  & $67.0_{(3.1)}$ \\
    & Re-weight     & $70.6_{(4.3)}$ & $69.9_{(5.1)}$  & $81.9_{(2.8)}$ & $60.8_{(1.8)}$ & $56.7_{(2.4)}$ & $75.4_{(1.3)}$ & $65.2_{(7.6)}$ & $65.0_{(8.2)}$ & $78.3_{(5.2)}$ & $57.9_{(4.3)}$  & $54.8_{(5.5)}$  & $73.3_{(3.0)}$ \\
    & SMOTE          & $65.1_{(4.0)}$ & $62.3_{(5.1)}$  & $78.3_{(2.7)}$ & $59.0_{(2.2)}$ & $53.9_{(2.6)}$ & $74.2_{(1.5)}$ & $60.3_{(7.6)}$ & $58.7_{(8.9)}$ & $74.9_{(5.3)}$ & $49.1_{(4.1)}$  & $45.6_{(5.4)}$  & $67.0_{(3.1)}$ \\
    & Embed-SMOTE    & $61.0_{(3.6)}$ & $58.0_{(5.5)}$  & $75.5_{(2.5)}$ & $55.5_{(2.1)}$ & $50.0_{(3.1)}$ & $71.7_{(1.5)}$ & $53.2_{(5.2)}$ & $50.9_{(6.4)}$ & $70.0_{(3.8)}$ & $40.7_{(2.8)}$  & $36.5_{(3.0)}$  & $60.5_{(2.2)}$ \\
    & GraphSMOTE     & $70.0_{(3.4)}$ & $68.6_{(4.9)}$  & $81.6_{(2.2)}$ & $62.5_{(1.8)}$ & $58.7_{(2.1)}$ & $76.5_{(1.2)}$ & $66.3_{(6.6)}$ & $65.3_{(7.7)}$ & $79.0_{(4.5)}$ & $55.8_{(5.6)}$  & $52.4_{(4.7)}$  & $71.9_{(4.0)}$ \\
    & GraphENS       & $59.3_{(7.0)}$ & $55.4_{(10.6)}$ & $74.2_{(4.9)}$ & $55.1_{(4.9)}$ & $48.1_{(7.9)}$ & $71.3_{(3.5)}$ & $44.3_{(6.5)}$ & $41.0_{(7.0)}$ & $63.3_{(5.0)}$ & $36.1_{(10.1)}$ & $31.1_{(12.3)}$ & $56.3_{(8.4)}$ \\
    & LTE4G          & $73.2_{(5.4)}$ & $72.1_{(6.1)}$  & $83.6_{(3.5)}$ & $70.9_{(2.5)}$ & $69.6_{(2.8)}$ & $82.1_{(1.6)}$ & $75.4_{(5.6)}$ & $75.4_{(5.4)}$ & $85.0_{(3.6)}$ & $70.2_{(4.5)}$  & $\mathbf{68.8}_{(4.7)}$  & $81.7_{(3.0)}$ \\
    & Uni-GNN     & $\mathbf{76.5}_{(0.5)}$ & $\mathbf{76.4}_{(0.7)}$  & $\mathbf{85.8}_{(0.3)}$ & $\mathbf{71.5}_{(1.2)}$ & $\mathbf{70.7}_{(1.5)}$ & $\mathbf{82.5}_{(0.8)}$ & $\mathbf{75.4}_{(3.7)}$ & $\mathbf{75.4}_{(3.7)}$ & $\mathbf{85.0}_{(2.4)}$ & $\mathbf{70.5}_{(3.7)}$  & ${68.7}_{(2.6)}$  & $\mathbf{81.8}_{(2.5)}$ \\

\hline
\hline
\multirow{9}{*}{\rotatebox{90}{CiteSeer}}  & GCN        & $49.5_{(2.1)}$ & $43.1_{(2.3)}$  & $66.7_{(1.5)}$ & $48.2_{(0.9)}$ & $39.3_{(0.4)}$ & $65.7_{(0.7)}$ & $48.9_{(1.4)}$ & $45.3_{(1.3)}$ & $66.2_{(1.1)}$ & $42.4_{(6.5)}$  & $39.1_{(7.3)}$  & $61.1_{(5.1)}$ \\
        & Over-sampling  & $51.5_{(3.0)}$ & $43.7_{(2.1)}$  & $68.2_{(2.2)}$ & $47.8_{(0.8)}$ & $38.9_{(1.9)}$ & $65.4_{(0.6)}$ & $43.0_{(3.4)}$ & $40.3_{(1.7)}$ & $61.7_{(2.7)}$ & $40.1_{(2.0)}$  & $34.2_{(1.5)}$  & $59.4_{(1.6)}$ \\
        & Re-weight      & $52.1_{(2.7)}$ & $46.2_{(3.2)}$  & $68.6_{(2.0)}$ & $48.0_{(0.4)}$ & $39.2_{(1.1)}$ & $65.6_{(0.3)}$ & $48.4_{(3.9)}$ & $44.5_{(3.9)}$ & $65.8_{(2.9)}$ & $41.3_{(4.5)}$  & $35.6_{(5.3)}$  & $60.3_{(3.6)}$ \\
        & SMOTE          & $48.7_{(2.5)}$ & $40.1_{(1.8)}$  & $66.1_{(1.9)}$ & $47.8_{(0.8)}$ & $38.9_{(1.9)}$ & $65.4_{(0.6)}$ & $44.9_{(4.4)}$ & $41.9_{(4.1)}$ & $63.2_{(3.4)}$ & $40.1_{(2.0)}$  & $34.2_{(1.5)}$  & $59.4_{(1.6)}$ \\
        & Embed-SMOTE    & $47.5_{(2.1)}$ & $37.9_{(1.7)}$  & $65.2_{(1.6)}$ & $46.7_{(3.0)}$ & $35.7_{(2.8)}$ & $64.5_{(2.3)}$ & $43.2_{(6.5)}$ & $38.3_{(5.8)}$ & $61.8_{(5.2)}$ & $33.2_{(6.6)}$  & $28.3_{(7.9)}$  & $53.4_{(5.9)}$ \\
        & GraphSMOTE    & $51.2_{(3.7)}$ & $43.4_{(4.2)}$  & $67.9_{(2.8)}$ & $49.3_{(2.0)}$ & $40.1_{(1.3)}$ & $66.5_{(1.5)}$ & $50.3_{(5.0)}$ & $46.1_{(4.5)}$ & $67.2_{(3.7)}$ & $46.5_{(3.7)}$  & $41.5_{(4.1)}$  & $64.4_{(2.9)}$ \\
        & GraphENS       & $44.2_{(3.5)}$ & $35.9_{(1.0)}$  & $62.7_{(2.7)}$ & $43.5_{(2.6)}$ & $33.4_{(1.9)}$ & $62.1_{(2.1)}$ & $33.0_{(3.2)}$ & $28.6_{(4.4)}$ & $53.4_{(2.9)}$ & $28.5_{(6.7)}$  & $23.1_{(6.2)}$  & $49.1_{(6.2)}$ \\
        & LTE4G          & $54.2_{(4.5)}$ & $51.8_{(4.1)}$  & $70.2_{(3.3)}$ & $52.7_{(2.1)}$ & $48.3_{(3.7)}$ & $69.1_{(1.5)}$ & $52.1_{(3.7)}$ & $47.2_{(3.6)}$ & $68.6_{(2.7)}$ & $47.3_{(1.1)}$  & $41.2_{(2.1)}$  & $65.0_{(0.9)}$ \\
        & Uni-GNN     & $\mathbf{59.1}_{(3.6)}$ & $\mathbf{54.6}_{(3.3)}$  & $\mathbf{73.6}_{(2.5)}$ & $\mathbf{54.1}_{(3.1)}$ & $\mathbf{48.5}_{(4.1)}$ & $\mathbf{70.1}_{(2.3)}$ & $\mathbf{58.3}_{(2.5)}$ & $\mathbf{55.0}_{(1.3)}$ & $\mathbf{73.1}_{(1.8)}$ & $\mathbf{54.0}_{(2.2)}$  & $\mathbf{51.4}_{(2.2)}$  & $\mathbf{70.0}_{(1.6)}$ \\ 
\hline                          
\hline

& \# Min. Class  & \multicolumn{6}{l|}{1}     & \multicolumn{6}{l}{2}               \\
\hline
  & $\rho$     & \multicolumn{3}{l|}{10\%}         & \multicolumn{3}{l|}{5\%}   & \multicolumn{3}{l|}{10\%}                          & \multicolumn{3}{l}{5\%}                           \\
\hline  
\multirow{8}{*}{\rotatebox{90}{PubMed}}  & GCN       & $60.4_{(6.5)}$ & $55.9_{(9.5)}$ & $69.6_{(5.2)}$ & $58.6_{(3.0)}$ & $51.9_{(6.2)}$ & $68.1_{(2.4)}$ & $49.1_{(10.9)}$ & $44.0_{(14.5)}$ & $60.3_{(8.9)}$ & $41.0_{(5.6)}$  & $32.2_{(8.4)}$  & $53.7_{(4.7)}$ \\
&  Oversampling & $57.6_{(0.5)}$ & $51.3_{(4.0)}$ & $67.4_{(0.4)}$ & $55.2_{(2.8)}$ & $46.8_{(2.2)}$ & $65.4_{(2.2)}$ & $41.6_{(5.7)}$  & $33.5_{(9.4)}$  & $54.2_{(4.8)}$ & $36.6_{(2.1)}$  & $23.8_{(4.8)}$  & $50.0_{(1.8)}$ \\
&  Re-weight     & $62.2_{(5.7)}$ & $57.6_{(9.2)}$ & $71.0_{(4.5)}$ & $59.4_{(3.6)}$ & $53.3_{(7.8)}$ & $68.8_{(2.8)}$ & $54.7_{(11.2)}$ & $53.8_{(11.7)}$ & $64.9_{(9.1)}$ & $47.5_{(11.4)}$ & $42.6_{(13.4)}$ & $59.0_{(9.5)}$ \\
&  SMOTE        & $55.8_{(3.0)}$ & $48.2_{(3.3)}$ & $65.9_{(2.4)}$ & $55.2_{(2.8)}$ & $46.8_{(2.2)}$ & $65.4_{(2.2)}$ & $41.8_{(4.0)}$  & $32.6_{(6.6)}$  & $54.4_{(3.3)}$ & $36.6_{(2.1)}$  & $23.8_{(4.8)}$  & $50.0_{(1.8)}$ \\
&  Embed-SMOTE & $53.4_{(1.8)}$ & $44.9_{(2.0)}$ & $63.9_{(1.4)}$ & $53.3_{(3.1)}$ & $43.4_{(3.4)}$ & $63.9_{(2.5)}$ & $38.6_{(1.2)}$  & $28.3_{(1.1)}$  & $51.7_{(1.1)}$ & $35.2_{(1.3)}$  & $21.4_{(3.5)}$  & $48.7_{(1.1)}$ \\
&  GraphSMOTE   & OOM            & OOM            & OOM            & OOM            & OOM            & OOM            & OOM             & OOM             & OOM            & OOM             & OOM             & OOM            \\
&  LTE4G        & $62.6_{(3.0)}$ & $59.2_{(6.7)}$ & $71.4_{(2.4)}$ & $60.0_{(5.4)}$ & $55.3_{(8.2)}$ & $69.3_{(4.3)}$ & $52.1_{(7.0)}$  & $50.2_{(8.3)}$  & $62.9_{(5.7)}$ & $48.5_{(9.9)}$  & $44.3_{(12.2)}$ & $48.5_{(9.9)}$ \\
&  Uni-GNN   &   $\mathbf{73.9}_{(2.3)}$ & $\mathbf{73.8}_{(2.5)}$ & $\mathbf{80.2}_{(1.8)}$ & $\mathbf{67.1}_{(4.2)}$ & $\mathbf{65.2}_{(5.3)}$ & $\mathbf{74.8}_{(3.3)}$ & $\mathbf{66.9}_{(4.3)}$  & $\mathbf{66.0}_{(4.1)}$  & $\mathbf{74.7}_{(3.3)}$ & $\mathbf{63.4}_{(6.4)}$  & $\mathbf{62.3}_{(8.6)}$  & $\mathbf{72.0}_{(5.0)}$ \\ 
\hline
\end{tabular}
}
\label{table:main_results}
\vskip -0.1in

\end{table*}

\subsection{Balanced Pseudo-Label Generation}
To leverage the unlabeled nodes in the graph, a balanced pseudo-label generation mechanism is proposed. The objective is to increase the number of available labeled nodes in the graph while considering the class imbalance in the set of labeled nodes. The goal is to generate more pseudo-labels from minority classes and fewer pseudo-labels from majority classes, thus balancing the class label distribution of the training data. 
In particular, for each class $c$, 
the number of nodes to pseudo-label, denoted as $\hat{N}_c$, 
is set as the difference between the largest labeled class size
and the size of class $c$, 
aiming to balance the class label distribution over the union of labeled nodes and pseudo-labeled nodes:
\begin{equation}
\hat{N}_c =\text{max}_{i \in \{1, \ldots, C\}}(|V^{i}_{l}|) - |V^c_l| 
\end{equation}
The set of unlabeled nodes that can be confidently pseudo-labeled 
to class $c$ can be determined as:  
\begin{equation}
\!\!\!	
	\hat{V}_u^c = \{i \mid \max(P_i) > \epsilon, \, \text{argmax}(P_i) = c, i \in V_u\}
\label{eq:pseudo_all}
\end{equation}
where $\epsilon$ is a hyperparameter determining the confidence prediction threshold.
Balanced sampling is then performed on each set $\hat{V}_u^c$ 
by selecting the top $\hat{N}_c$ nodes, 
denoted as $\text{Top}_{\hat{N}_c}(\hat{V}_u^c)$,  
with the most confident pseudo-labels based on the predicted probability $P_i[c]$.  
This results in a total set of 
pseudo-labeled nodes, denoted as $\hat{V}_u$, from all classes:
\begin{equation}
\hat{V}_u =  \{ \text{Top}_{\hat{N}_1}( \hat{V}_u^1), \cdots, \text{Top}_{\hat{N}_C}( \hat{V}_u^C)  \}. 
\label{eq:pseudo_select}
\end{equation}

The Unified GNN Learning framework is trained to minimize the following node classification loss 
over this set of pseudo-labeled nodes $\hat{V}_u$:
\begin{equation}
	\mathcal{L}_{\text{ps}} = \frac{1}{|\hat{V}_u|} \sum\nolimits_{i \in \hat{V}_u } \ell_{\text{ce}}( P_i , 
	\boldsymbol{1}_{\text{argmax}(P_i)}  ) 
\label{eq:ps_loss}
\end{equation}
where $\ell_{\text{ce}}$ again is the standard cross-entropy loss function, 
$P_i$ is the predicted class probability vector with classifier $\phi$, 
and 
$\boldsymbol{1}_{\text{argmax}(P_i)}$ is a one-hot pseudo-label vector
with a single 1 at the predicted class entry $\text{argmax}(P_i)$. 
This pseudo-labeling mechanism aims to augment the labeled node set, particularly focusing on addressing class imbalances 
by generating more pseudo-labels for minority classes.

\paragraph{Training Loss}
The unified GNN Learning framework is trained on the labeled set $V_{l}$ 
and the selected pseudo-labeled set $\hat{V}_u$ in an end-to-end fashion to minimize the following integrated total loss: 
\begin{equation}
\mathcal{L}_{\text{total}} = \mathcal{L}  + \lambda  \mathcal{L}_{\text{ps}}
\label{eq:total_loss}
\end{equation}
where $\lambda$ is a hyper-parameter. 
The training procedure is provided in the Appendix.

\section{Experiments}

\subsection{Experimental Setup}

\paragraph{Datasets \& Baselines}
We conduct experiments on three
datasets (Cora, CiteSeer and PubMed) \cite{sen2008collective}. 
To ensure a fair comparison, we adhere to the evaluation protocol used in previous studies \cite{zhao2021graphsmote,yun2022lte4g}. The datasets undergo manual pre-processing to achieve the desired imbalance ratio ($\rho$).
Specifically, each majority class is allocated 20 labeled training nodes, while each minority class is assigned $20\times\rho$ labeled training nodes. For validation and test sets, each class is assigned 25 and 55 nodes, respectively.
Following the protocol of \cite{yun2022lte4g}, we consider two imbalance ratios 
($\rho=10\%, 5\%$), and two different numbers of minority classes:
3 and 5 on Cora and CiteSeer, and 1 and 2 on the PubMed dataset. Additionally, for Cora and CiteSeer, 
we also adopt a long-tail class label distribution setup. Here, low-degree nodes from minority classes are removed from the graph until the desired imbalance ratio ($\rho$) of 1\% and class label distribution are achieved, similar to \cite{park2021graphens,yun2022lte4g}. 
We compare the proposed Uni-GNN with
the underlying GCN baseline, as well as various traditional and graph-based class-imbalanced learning baselines: Over-sampling, Re-weight \cite{yuan2012sampling}, SMOTE \cite{chawla2002smote}, Embed-SMOTE \cite{ando2017deep}, GraphSMOTE \cite{zhao2021graphsmote}, GraphENS \cite{park2021graphens}, and LTE4G \cite{yun2022lte4g}.

\noindent{\bf Implementation Details}\quad
Graph Convolution Network (GCN) \cite{kipf2017semisupervised} 
implements the message passing layers in our proposed framework and all the comparison baselines.
The semantic and structural encoders consist of 2 message passing layers each, followed by a 
ReLU activation function. The node classifier is composed of a single GCN 
layer, followed by ReLU activation, and then a single fully connected linear layer.
Uni-GNN undergoes training using an Adam optimizer with a learning rate of $1e^{-2}$ and weight decay of $5e^{-4}$ over 10,000 epochs. We incorporate an early stopping criterion with a patience of 1,000 epochs and apply a dropout rate of 0.5 to all layers of our framework. 
The size of the learned hidden embeddings for all layers of structural and semantic encoders is set to 64. The hyperparameter $\lambda$ is assigned the value 1.
For the hyperparameters $K$, $\alpha$, $\epsilon$, and $\beta$, we explore the following ranges: $\{3C, 4C, 10C, 20C, 30C\}$, $\{1, 2\}$, $\{0.5, 0.7\}$, and $\{50, 100\}$, respectively. Each experiment is repeated three times, and the reported performance metrics represent the mean and standard deviation across all three runs.
For results for comparison methods on Cora and CiteSeer, 
we refer to the outcomes from \cite{yun2022lte4g}.

\begin{table}[t]
\centering
\caption{The overall performance (standard deviation within brackets) on Cora-LT and CiteSeer-LT datasets with $\rho=1\%$.}
\setlength{\tabcolsep}{1.2pt}
\resizebox{ 0.48 \textwidth}{!}{
\begin{tabular}{l|lll|lll}
\hline
              & \multicolumn{3}{l|}{Cora-LT}                       & \multicolumn{3}{l}{CiteSeer-LT}                  \\
\hline              
              & bAcc.           & Macro-F1       & G-Means        & bAcc.          & Macro-F1       & G-Means        \\
\hline              
GCN        & $66.8_{(1.1)}$  & $65.0_{(1.0)}$ & $79.5_{(0.7)}$ & $50.4_{(1.4)}$ & $45.7_{(0.8)}$ & $67.4_{(1.1)}$ \\
Over-sampling & $66.6_{(0.8)}$  & $64.5_{(0.6)}$ & $79.3_{(0.5)}$ & $51.7_{(1.2)}$ & $46.7_{(0.8)}$ & $68.4_{(0.9)}$ \\
Re-weight     & $68.0_{(0.7)}$  & $66.7_{(1.4)}$ & $80.2_{(0.5)}$ & $53.0_{(2.4)}$ & $48.8_{(2.3)}$ & $69.3_{(1.7)}$ \\
SMOTE         & $66.8_{(0.4)}$  & $66.4_{(0.6)}$ & $79.4_{(0.2)}$ & $51.2_{(1.9)}$ & $46.6_{(1.8)}$ & $68.0_{(1.4)}$ \\
Embed-SMOTE   & $65.2_{(0.6)}$  & $63.1_{(0.3)}$ & $78.4_{(0.4)}$ & $52.6_{(1.6)}$ & $48.0_{(1.5)}$ & $69.0_{(1.2)}$ \\
GraphSMOTE    & $67.7_{(1.2)}$  & $66.3_{(1.1)}$ & $80.0_{(0.8)}$ & $51.5_{(0.7)}$ & $46.8_{(0.9)}$ & $68.2_{(0.5)}$ \\
GraphENS      & $67.4_{(1.5)}$  & $64.5_{(1.5)}$ & $79.8_{(1.0)}$ & $52.4_{(1.7)}$ & $47.2_{(1.0)}$ & $68.9_{(1.3)}$ \\
LTE4G         & $72.2_{(3.1)}$  & $72.0_{(2.9)}$ & $83.0_{(2.0)}$ & $56.4_{(2.1)}$ & $52.5_{(2.2)}$ & $71.7_{(1.5)}$ \\
Uni-GNN    & $\mathbf{75.2}_{(1.3 )}$ & $\mathbf{74.8}_{(1.3)}$ & $\mathbf{84.9}_{(0.8)}$ & $\mathbf{63.3}_{(1.7)}$ & $\mathbf{61.6}_{(2.5)}$ & $\mathbf{76.6}_{(1.2)}$ \\ 
\hline
\end{tabular}}
\label{table:LT_results}
\vskip -0.1in

\end{table}

\subsection{Comparison Results}

\begin{table*}[t]
\caption{Ablation study results (standard deviation within brackets) on Cora and CiteSeer datasets with 3 minority classes and 10\% imbalance ratio and Long-Tail class label distribution with 1\% imbalance ratio.}
\setlength{\tabcolsep}{3.0pt}
\resizebox{\textwidth}{!}{ 
\begin{tabular}{l|lll|lll|lll|lll}
\hline
(\# Min. Class, $\rho$ ) & \multicolumn{3}{l|}{Cora (3, 10\%)} & \multicolumn{3}{l|}{CiteSeer (3, 10\%)} & \multicolumn{3}{l|}{Cora (LT, 1\%)} & \multicolumn{3}{l}{CiteSeer (LT, 1\%)} \\
\hline
     & bAcc.           & Macro-F1       & G-Means        & bAcc.          & Macro-F1       & G-Means        & bAcc.           & Macro-F1       & G-Means        & bAcc.           & Macro-F1       & G-Means        \\

\hline
GCN         & ${68.8}_{(4.0)}$         & ${67.6}_{(5.0)}$          & ${80.8}_{(2.6)}$   & ${49.5}_{(2.1)}$          & ${43.1}_{(2.3)}$          & ${66.7}_{(1.5)}$     &     $66.8_{(1.1)}$  &   $65.0_{(1.0)}$  &   $79.5_{(0.7)}$  &   $50.4_{(1.4)}$  &	$45.7_{(0.8)}$    &   $67.4_{(1.1)}$    \\

Uni-GNN  & $\mathbf{76.5}_{(0.5)}$ & $\mathbf{76.4}_{(0.7)}$  & $\mathbf{85.8}_{(0.3)}$  & $\mathbf{59.1}_{(3.6)}$ & $\mathbf{54.6}_{(3.3)}$  & $\mathbf{73.6}_{(2.5)}$    &   $\mathbf{75.2}_{(1.3)}$  & 	$\mathbf{74.8}_{(1.3)}$   &    $\mathbf{84.9}_{(0.8)}$   &  $\mathbf{63.3}_{(1.7)}$  &   $\mathbf{61.6}_{(2.5)}$ &   $\mathbf{76.6}_{(1.2)}$\\
$\; -$ Ind. Enc.  & ${74.3}_{(3.9)}$          & ${74.0}_{(4.2)}$          & ${84.3}_{(2.5)}$          & ${54.2}_{(5.0)}$          & ${51.7}_{(5.4)}$          & ${70.1}_{(3.6)}$   &   $74.9_{(1.0)}$  &   $74.3_{(0.8)}$  &   $84.7_{(0.6)}$  &   $60.4_{(0.9)}$  &   $58.2_{(1.5)}$  &   $74.6_{(0.7)}$  \\
$\; - $ Drop $\mathcal{L}_{\text{ps}}$  & ${72.6}_{(1.5)}$          & ${72.3}_{(1.6)}$          & ${83.3}_{(1.0)}$          & ${57.3}_{(4.6)}$          & ${53.8}_{(4.4)}$          & ${72.3}_{(3.2)}$    & $74.1_{(1.2)}$    &   $73.6_{(1.2)}$  &   $84.2_{(0.8)}$   &   $57.7_{(3.9)}$  &   $53.8_{(4.0)}$  &   $72.6_{(2.8)}$   \\

$\; - \, A_{\text{struct}}= A$  & ${74.9}_{(1.5)}$          & ${75.2}_{(1.2)}$          & ${84.7}_{(1.0)}$          & ${56.0}_{(2.5)}$          & ${52.3}_{(2.8)}$          & ${71.4}_{(1.8)}$    &   ${75.2}_{(1.3)}$  & 	${74.8}_{(1.3)}$   &    ${84.9}_{(0.8)}$   &  ${63.3}_{(1.7)}$  &   ${61.6}_{(2.5)}$ &   ${76.6}_{(1.2)}$\\

$\; - $ Semantic Enc. & ${61.9}_{(0.7)}$          & ${61.9}_{(0.7)}$          & ${76.1}_{(0.5)}$          & ${46.6}_{(1.4)}$        & ${42.6}_{(1.4)}$          & ${64.5}_{(1.0)}$   &   $57.6_{(3.2)}$  &   $56.3_{(3.6)}$  &   $73.1_{(2.2)}$  &   $51.3_{(2.9)}$  &   $49.2_{(3.2)}$  &   $68.0_{(2.1)}$   \\
$\; - $ Structural Enc. & ${73.6}_{(3.3)}$          & ${73.0}_{(3.5)}$          & ${83.9}_{(2.1)}$          & ${53.3}_{(2.6)}$          & ${49.2}_{(3.5)}$          & ${69.5}_{(1.9)}$   &   $73.2_{(0.9)}$  &   $72.3_{(0.7)}$  &   $83.7_{(0.6)}$  &   $61.2_{(1.9)}$  &   $59.0_{(1.9)}$  &   $75.1_{(1.3)}$\\

$\; -$ Imbalanced PL  & ${72.6}_{(1.8)}$          & ${72.1}_{(1.9)}$          & ${83.3}_{(1.2)}$          & ${46.9}_{(2.4)}$          & ${39.3}_{(3.0)}$          & ${64.7}_{(1.8)}$   &   $74.5_{(0.1)}$  &   $74.0_{(0.2)}$  &   $84.4_{(0.1)}$  &   $54.4_{(2.4)}$  &   $50.0_{(2.6)}$  &   $70.3_{(1.7)}$  \\

$\; - $ Fixed $\{A^{\ell}_{\text{sem}}\}_{\ell=2}^{L}$ & ${76.0}_{(2.5)}$          & ${75.9}_{(2.6)}$          & ${85.4}_{(1.6)}$          & ${55.8}_{(2.9)}$          & ${52.8}_{(2.0)}$          & ${71.3}_{(2.0)}$   &   $74.4_{(0.2)}$ &   $73.5_{(0.4)}$  &   $84.4_{(0.2)}$  &   $61.9_{(1.5)}$  &   $60.4_{(0.9)}$  &   $75.6_{(1.0)}$   \\

$\; - \, \, \omega_{c_i} = 1, \forall i \in V_{l}$   & ${71.1}_{(3.1)}$          & ${70.7}_{(3.4)}$          & ${82.2}_{(2.0)}$          & ${51.7}_{(4.2)}$          & ${48.6}_{(4.7)}$          & ${68.3}_{(3.0)}$    &   $57.0_{(1.9)}$  &   $48.7_{(1.3)}$  &   $72.7_{(1.4)}$  &   $57.9_{(1.9)}$  &   $52.8_{(2.1)}$  &   $72.8_{(1.3)}$  \\

\hline
\end{tabular}}
\label{table:ablation}

\end{table*}

\begin{figure*}[t]
\centering
\begin{subfigure}{0.24\textwidth}
\centering
\includegraphics[width = \textwidth]{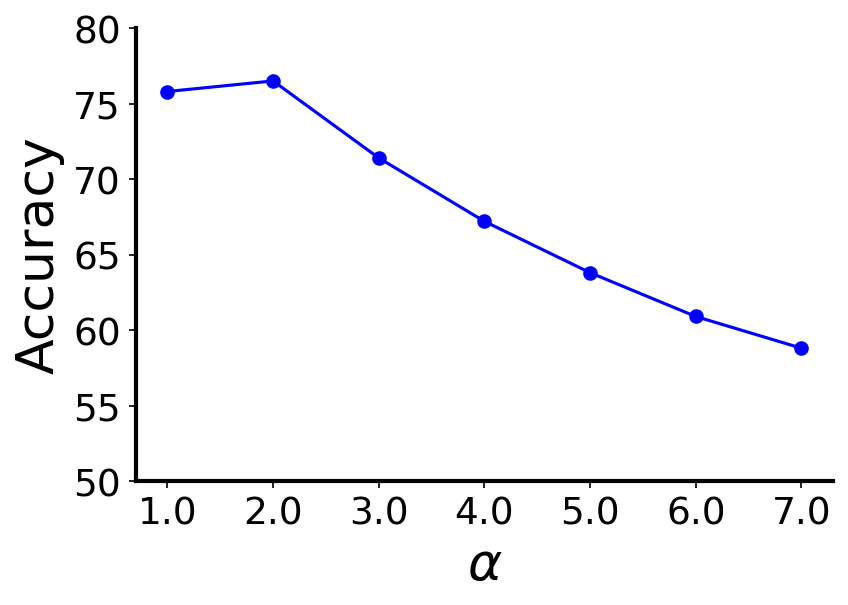} 
\caption{$\alpha$}
\label{fig:sen_a}
\end{subfigure}
\begin{subfigure}{0.24\textwidth}
\centering
\includegraphics[width = \textwidth]{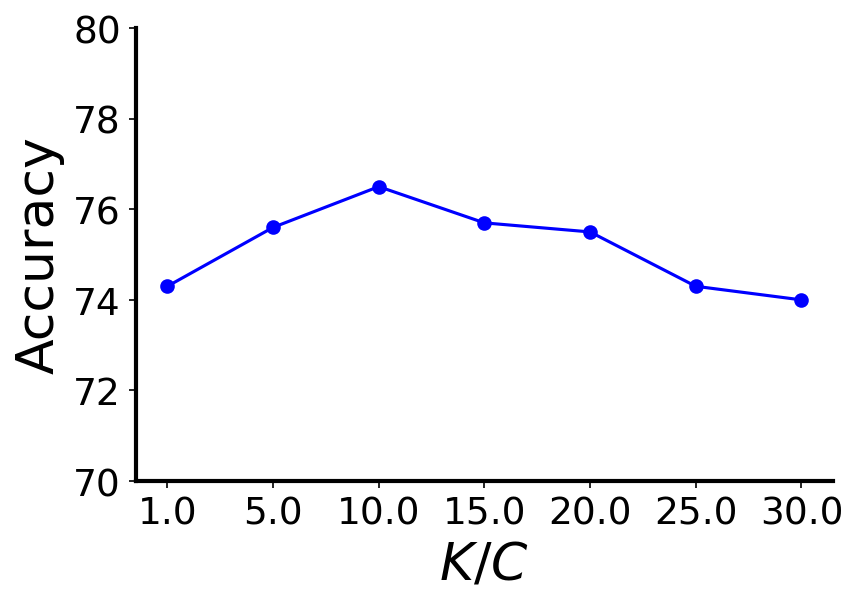}
\caption{$K$}
\label{fig:sen_b}
\end{subfigure}
\begin{subfigure}{0.24\textwidth}
\centering
\includegraphics[width = \textwidth]{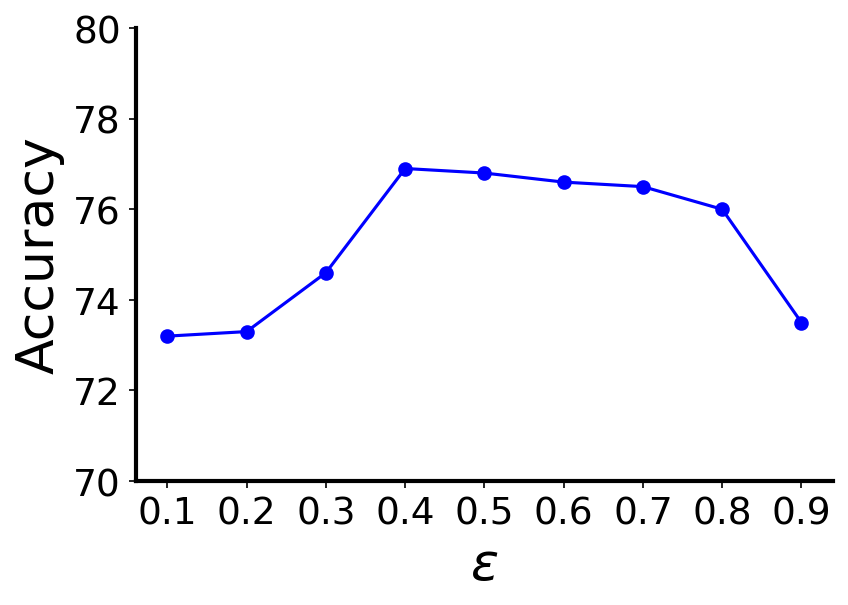}
\caption{$\epsilon$}
\label{fig:sen_c}
\end{subfigure}
\begin{subfigure}{0.24\textwidth}
\centering
\includegraphics[width = \textwidth]{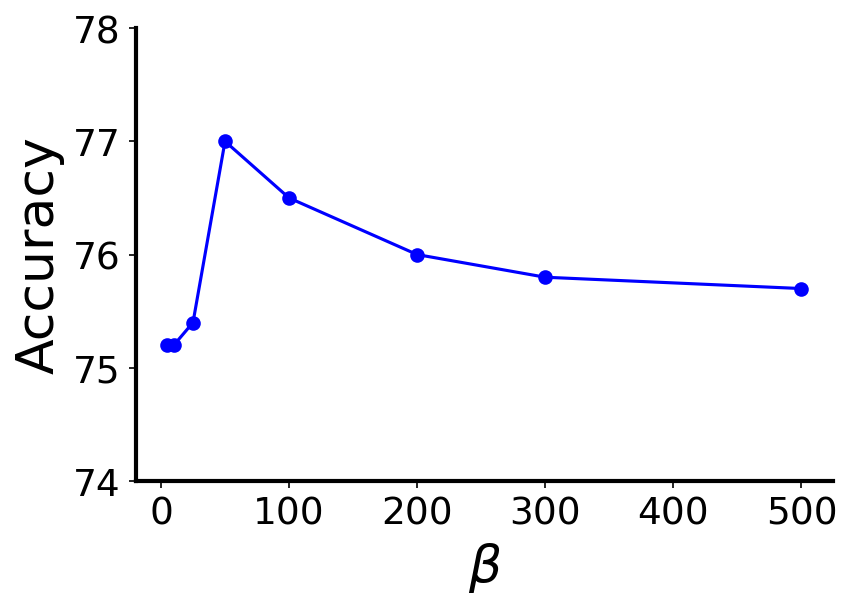} 
\caption{$\beta$}
\label{fig:sen_d}

\end{subfigure}
\caption{Sensitivity analysis for the proposed framework on hyper-parameters (a) $\alpha$, the max SPD distance in $A_{\text{struct}}$; (b) $K$, the number of clusters; (c) $\epsilon$, the pseudo-label confidence threshold; (d) $\beta$, the rate of updating $\{A^{\ell}_{\text{sem}}\}_{\ell=2}^{L}$.}
\label{fig:hyper_sen}
\vskip -0.1in
\end{figure*}

We evaluate the performance of
Uni-GNN framework on the semi-supervised node classification task under class imbalance. Across the three datasets, we explore four distinct evaluation setups by manipulating the number of minority classes and the imbalance ratio ($\rho$). 
Additionally, we explore the long-tail class label distribution setting for Cora and CiteSeer with an imbalance ratio of $\rho = 1\%$.
We assess the performance of Uni-GNN using balanced Accuracy (bAcc), Macro-F1, and Geometric Means (G-Means), reporting the mean and standard deviation of each metric over 3 runs. Table \ref{table:main_results} summarizes the results for different numbers of minority classes and imbalance ratios on all three datasets, while Table \ref{table:LT_results} showcases the results under long-tail class label distribution on Cora and CiteSeer.

Table \ref{table:main_results} 
illustrates that the performance of all methods diminishes with decreasing imbalance ratio ($\rho$) and increasing numbers of minority classes. Our proposed framework consistently outperforms the underlying GCN baseline and all other methods across all three datasets and various evaluation setups. The performance gains over the GCN baseline are substantial, exceeding 10\% in most cases for Cora and CiteSeer datasets and 13\% for most instances of the PubMed dataset.
Moreover, Uni-GNN consistently demonstrates superior performance compared to all other comparison methods, achieving notable improvements over the second-best method (LTE4G) by around 3\%, 5\%, and 11\% on Cora, CiteSeer with 3 minority classes and $\rho=10\%$, and PubMed with 1 minority class and $\rho=10\%$, respectively.
Similarly, Table \ref{table:LT_results} highlights that Uni-GNN 
consistently enhances the performance of the underlying GCN baseline, achieving performance gains exceeding 8\% and 12\% on Cora-LT and CiteSeer-LT datasets, respectively. Furthermore, Uni-GNN demonstrates remarkable performance gains over all other class imbalance methods, surpassing 3\% and 6\% on Cora-LT and CiteSeer-LT, respectively. These results underscore the superior performance of our 
framework over existing state-of-the-art class-imbalanced GNN methods across numerous challenging class imbalance scenarios.

\subsection{Ablation Study}

We conducted an ablation study to discern the individual contributions of each component in our proposed framework. In addition to the underlying GCN baseline, eight variants were considered: 
(1) Independent Node Encoders (Ind. Enc.): 
each node encoder exclusively propagates its own node embeddings, instead of propagating the concatenated semantic and structural embeddings. 
Specifically, $f^{\ell}_{\text{sem}}$ solely propagates ${H}_{\text{sem}}^{\ell-1}$ while $f^{\ell}_{\text{struct}}$ solely propagates ${H}_{\text{struct}}^{\ell-1}$. 
(2) Drop $\mathcal{L}_{\text{ps}}$: excluding the balanced pseudo-label generation. 
(3) $A_{\text{struct}}= A$: structural connectivity considers only immediate neighbors of nodes ($\alpha=1$). 
(4) Semantic Encoder only (Semantic Enc.): it discards the structural encoder.
(5) Structural Encoder only (Structural Enc.): it discards the semantic encoder.
(6) Imbalanced Pseudo-Labeling: it generates pseudo-labels for all unlabeled nodes with confident predictions without considering class imbalance. 
(7) Fixed $\{A^{\ell}_{\text{sem}}\}_{\ell=2}^{L}$: it does not update the semantic connectivity during training. 
(8) $\omega_{c_i} = 1, \forall i \in V_{l}$: it assigns equal weights to all labeled nodes in the training set. Evaluation was performed on Cora and CiteSeer datasets, each with 3 minority classes and imbalance ratio ($\rho$) of 10\%, 
and with long-tail class label distribution and imbalance ratio ($\rho$) of 1\%. The results are reported in Table \ref{table:ablation}.

Table \ref{table:ablation} illustrates performance degradation in all variants compared to the full proposed framework. 
The observed performance decline in the Independent Node Encoders (Ind. Enc.) variant underscores the importance of simultaneously propagating semantic and structural embeddings across both the semantic and structural connectivity. This emphasizes the need for incorporating both aspects to effectively learn more discriminative node embeddings.
The performance drop observed in the Semantic Enc. and Structural Enc. variants underscores the significance and individual contribution of each connectivity type to the proposed Uni-GNN framework. This highlights the critical role that each type of connectivity plays in the overall performance of the proposed framework. The $A_{\text{struct}}= A$ variant's performance drop emphasizes the importance of connecting nodes beyond their immediate structural neighbors, enabling the propagation of messages across a larger portion of the graph and learning more discriminative embeddings. The performance degradation in Drop $\mathcal{L}_{\text{ps}}$ and Imbalanced Pseudo-Labeling variants validates the substantial contribution of our balanced pseudo-label generation mechanism. The $\omega_{c_i} = 1, \forall i \in V_{l}$ variant underscores the importance of assigning weights to each node in the labeled training set based on their class frequency. The consistent performance drops across both datasets for all variants affirm the essential contribution of each corresponding component in the proposed framework.

\subsection{Hyper-Parameter Sensitivity}

To investigate the impact of the hyper-parameters in Uni-GNN framework, we present the results of several sensitivity analyses on the Cora dataset with 3 minority classes and an imbalance ratio of 10\% in Figure \ref{fig:hyper_sen}. Figures \ref{fig:sen_a}, \ref{fig:sen_b}, \ref{fig:sen_c}, and \ref{fig:sen_d} depict the accuracy of 
Uni-GNN as we independently vary
the max SPD distance in $A_{\text{struct}}$ ($\alpha$), the number of clusters ($K$), the pseudo-label confidence threshold ($\epsilon$) and the rate of updating $\{A^{\ell}_{\text{sem}}\}_{\ell=2}^{L}$ ($\beta$), respectively.
Larger values for $\alpha$ result in performance degradation due to over-smoothing as the graph becomes more densely connected. Optimal performance is achieved with $\alpha=2$.
Uni-GNN exhibits robustness to variations in the hyperparameters $K$, $\epsilon$, and $\beta$ within a broad range. It consistently outperforms state-of-the-art methods across diverse settings of these hyperparameters, as depicted by the corresponding results presented in Table \ref{table:main_results}.
Small $K$ values lead to noisy clusters with mixed-class nodes, while large $K$ values result in over-segmented clusters with sparse semantic connectivity. Optimal performance is achieved when $K$ falls within the range of $5C$ to $15C$. 
Inadequately small values for $\epsilon$ result in the utilization of noisy pseudo-label predictions in the training process, while excessively large values exclude reasonably confident pseudo-labeled nodes from selection. The optimal range for $\epsilon$ lies between 0.4 and 0.7.
Extremely small values for $\beta$ lead to frequent updates of the semantic connectivity, preventing Uni-GNN from learning stable discriminative embeddings. Values around 100 training epochs yield the best results.

\section{Conclusion}

In this paper, we introduced a novel 
Uni-GNN framework  for class-imbalanced node classification.
The proposed framework harnesses the 
combined strength
of structural and semantic connectivity through dedicated structural and semantic node encoders, enabling the learning of a unified node representation. 
By utilizing these encoders, the structural and semantic connectivity ensures effective
propagation of messages well beyond the 
structural immediate neighbors of nodes, 
thereby addressing the under-reaching and neighborhood memorization problems.
Moreover, we proposed a balanced pseudo-label generation mechanism 
to incorporate
confident pseudo-label predictions from minority unlabeled nodes into the training set.
Our experimental evaluations on three 
benchmark datasets for node classification affirm
the efficacy of our proposed framework. The results demonstrate that Uni-GNN 
adeptly mitigates class imbalance bias, surpassing existing state-of-the-art methods in 
class-imbalanced graph learning.

\bibliography{main}
\bibliographystyle{icml2024}
\newpage
\appendix
\onecolumn
\include{appendix}

\end{document}

%% file: appendix.tex
\section{Training Procedure}
The details of the training procedure of the Unified GNN Learning framework 
are presented in algorithm \ref{alg:method}.

\begin{algorithm}[h]
    \caption{Training Procedure for Uni-GNN Framework}
    \label{alg:method}
    \begin{algorithmic}[1]
\STATE{\textbf{Input:} Graph $G$ with input feature matrix $X$, adjacency matrix $A$, label indicator matrix $Y^{l}$ 
\\ Hyper-parameters $\alpha$, $\beta$, $\lambda$, $\epsilon$, $K$ }
\STATE{\textbf{Output:} Learned model parameters $\mathcal{W}_{\text{sem}}$, $\mathcal{W}_{\text{struct}}$, $\mathcal{W}_{\phi}$ }
        
\STATE Construct $A_{\text{struct}}$ using Eq.(1)
\STATE Construct $\{A^{\ell}_{\text{sem}}\}_{\ell=1}^{L}$ using Eq.(3), Eq.(4), Eq.(5)
\FOR{{iter = 1} {\bf to} maxiters}
\STATE ${H}_{\text{sem}}^{1} =  f^{1}_{\text{sem}}(X , A^{1}_{\text{sem}})$
\STATE ${H}_{\text{struct}}^{1} =  f^{1}_{\text{struct}}(X , A_{\text{struct}})$
\FOR{{$\ell$ = 2} {\bf to} L}
    \STATE ${H}_{\text{sem}}^{\ell} =  f^{\ell}_{\text{sem}}({H}_{\text{struct}}^{\ell-1} || {H}_{\text{sem}}^{\ell-1}  , A^{\ell}_{\text{sem}})$
    \STATE ${H}_{\text{struct}}^{\ell} =  f^{\ell}_{\text{struct}}({H}_{\text{struct}}^{\ell-1} || {H}_{\text{sem}}^{\ell-1}  , A_{\text{struct}})$
\ENDFOR

\STATE $P = \phi({H}_{\text{struct}}^{L} || {H}_{\text{sem}}^{L})$

\STATE Update $\hat{V}_u$ using Eq.(10), Eq.(11), Eq.(12)

\STATE Calculate $\mathcal{L}_{\text{total}}$ using Eq.(8), Eq.(13), Eq.(14)
\STATE{Update $\mathcal{W}_{\text{sem}}$, $\mathcal{W}_{\text{struct}}$, $\mathcal{W}_{\phi}$ to minimize $\mathcal{L}_{\text{total}}$ with gradient descent}

    \IF{$iter \mod \beta = 0$}  
\STATE Update $\{A^{\ell}_{\text{sem}}\}_{\ell=2}^{L}$ using Eq.(3), Eq.(4), Eq.(5)
    \ENDIF
    \ENDFOR
    \end{algorithmic}
\end{algorithm}

\section{Datasets}

The specifics of the benchmark datasets are detailed in Table \ref{table:dataset}. Detailed class label distribution information for the training sets in all evaluation setups on all datasets are provided in Table \ref{table:dataset_setups}.

\begin{table}[!h]
\centering
\caption{Details of the benchmark datasets.}
\setlength{\tabcolsep}{3.0pt}
\begin{tabular}{l|r|r|r|r}
\hline
Dataset  & \# Nodes & \# Edges & \# Features & \# Classes \\
\hline
Cora     & 2,708    & 5,278    & 1,433       & 7          \\
CiteSeer & 3,327    & 4,552    & 3,703       & 6          \\
PubMed   & 19,717   & 44,324   & 500         & 3         \\
\hline
\end{tabular}
\label{table:dataset}
\end{table}

\begin{table}[!h]
\centering
\caption{Details of the class label distribution of the training set of all evaluation setups on all datasets.}
\setlength{\tabcolsep}{3.0pt}
\resizebox{0.49\textwidth}{!}{
\begin{tabular}{l|l|r|ccccccc}
\hline
Dataset  & \# Min. Class      &   \multicolumn{1}{l|}{$\rho$}    & $|V^1_{l}|$ & $|V^2_{l}|$ & $|V^3_{l}|$ &$|V^4_{l}|$ &$|V^5_{l}|$ & $|V^6_{l}|$ & $|V^7_{l}|$\\
\hline
\multirow{5}{*}{Cora}     & \multirow{2}{*}{3} & 10\%   & 20  & 20 & 20  & 20 & 2  & 2  & 2  \\
    &  & 5\%   & 20  & 20  & 20    & 20  & 1 & 1   & 1   \\
\cline{2-10} & \multirow{2}{*}{5} & 10\%  & 20 & 20 & 2 & 2 & 2 & 2 & 2  \\
    &  & 5\%  & 20 & 20 & 1 & 1 & 1 & 1  & 1   \\
\cline{2-10} & LT   & 1\%  & 341 & 158 & 73 & 34  & 15 & 7  & 3   \\
\hline         
\multirow{5}{*}{CiteSeer} & \multirow{2}{*}{3} & 10\%   & 20  & 20  & 20    & 2  & 2  & 2 & -  \\
    &   & 5\%  & 20  & 20  & 20  & 1   & 1   & 1   & -   \\ 
\cline{2-10} 
    & \multirow{2}{*}{5} & 10\%  & 20  & 2  & 2  & 2 & 2 & 2 & -  \\
    &  & 5\%  & 20 & 1 & 1 & 1  & 1  & 1  & -  \\
\cline{2-10} 
    & LT & 1\%  & 371  & 147 & 58  & 32 & 9  & 3   & - \\
\hline         
\multirow{4}{*}{PubMed} & \multirow{2}{*}{1} & 10\%  & 20 & 20 & 2 & - & - & -  & -     \\
         &  & 5\%  & 20 & 20  & 1 & - & - & -  & - \\ 
\cline{2-10} 
   & \multirow{2}{*}{2} & 10\%    & 20  & 2   & 2   & - & - & -  & -     \\
   &         & 5\%    & 20   & 1   & 1   & - & -  & - & -   \\
\cline{2-10} 
\hline
\end{tabular}}
\label{table:dataset_setups}
\end{table}